# PLOT-CT: Pre-log Voronoi Decomposition Assisted Generation for Low-dose CT Reconstruction

Bin Huang, Xun Yu, Yikun Zhang, Yi Zhang, *Senior Member, IEEE*, Yang Chen, *Senior Member, IEEE*, Qiegen Liu, *Senior Member, IEEE*

*Abstract*—Low-dose computed tomography (LDCT) reconstruction is fundamentally challenged by severe noise and compromised data fidelity under reduced radiation exposure. Most existing methods operate either in the image or post-log projection domain, which fails to fully exploit the rich structural information in pre-log measurements while being highly susceptible to noise. The requisite logarithmic transformation critically amplifies noise within these data, imposing exceptional demands on reconstruction precision. To overcome these challenges, we propose PLOT-CT, a novel framework for Pre-Log vOronoi decomposiTion-assisted CT generation. Our method begins by applying Voronoi decomposition to pre-log sinograms, disentangling the data into distinct underlying components, which are embedded in separate latent spaces. This explicit decomposition significantly enhances the model's capacity to learn discriminative features, directly improving reconstruction accuracy by mitigating noise and preserving information inherent in the pre-log domain. Extensive experiments demonstrate that PLOT-CT achieves state-of-the-art performance, attaining a 2.36 dB PSNR improvement over traditional methods at the 1e4 incident photon level in the pre-log domain.

*Index Terms*—Low-dose CT reconstruction, pre-log domain, diffusion transformer, sinogram decomposition, Voronoi decomposition.

## I. INTRODUCTION

Computed tomography plays an indispensable role in modern clinical practice [1]. However, the growing imperative to reduce radiation exposure has driven the widespread adoption of LDCT, which inevitably introduces severe noise and reconstruction artifacts that substantially degrade diagnostic reliability [2-3]. Restoring image quality under low-dose conditions has therefore become a central challenge in CT imaging and a major focus of methodological research [4-6].

As illustrated in Fig. 1, the LDCT reconstruction process can be systematically interpreted as a sequence of three data domains, namely the image domain [7], the post-log projection domain [8], and the pre-log projection domain [9]. This domain-wise formulation reflects a fundamental trade-off between information accessibility and reconstruction difficulty. Moving upstream along the reconstruction pipeline provides access to increasingly raw and statistically informative measurement data, which are more amenable to physically grounded modeling and correction [10]. At the same time, operating closer to the data acquisition stage imposes increasingly stringent accuracy requirements, since restoration noises introduced early in the pipeline are magnified through subsequent non-linear transformations and reconstruction procedures, ultimately corrupting the final image [11].

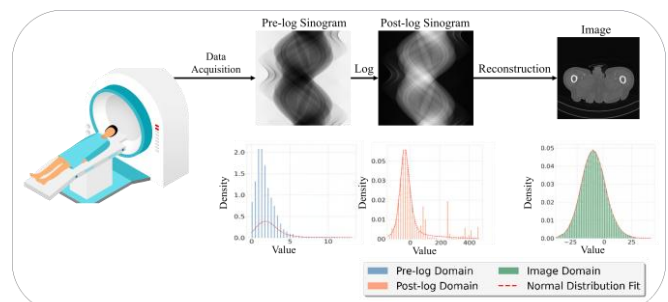

Fig. 1. Overview of CT reconstruction pipeline and corresponding probability density distribution plots of noises in the pre-log, post-log and image domains.

Image-domain methods perform processing directly on reconstructed CT slices and represent the most straightforward and computationally efficient reconstruction strategy [12-13]. Owing to the wide availability and standardized nature of reconstructed images in clinical workflows, image-domain data are generally easier to obtain and deploy than projection-level measurements [14]. Early deep learning approaches, including the residual encoder–decoder convolutional neural network introduced by Chen *et al.* [15] and the unrolled optimization framework with learned regularizers proposed by Ding *et al.* [16], have demonstrated substantial effectiveness in suppressing local noise and improving perceptual image quality. Nevertheless, image-domain processing is fundamentally constrained by the nature of reconstructed images, in which measurement noise and reconstruction artifacts have already been inseparably fused with anatomical structures during the reconstruction process [17]. In the absence of access to projection-level measurements, correction is limited to post hoc image refinement rather than physically

This work was supported by National Natural Science Foundation of China (Grant number 62122033, 62201193), and in part by Nanchang University Interdisciplinary Innovation Fund Project (Grant number PYJX20230002). (B. Huang and X. Yu are co-first authors.) (Corresponding author: Q. Liu.)

B. Huang and X. Yu are with School of Mathematics and Computer Sciences, Nanchang University, Nanchang, China ({huangbin, 409100240062}@email.ncu.edu.cn).
Y. Zhang and Y. Chen are with School of Computer Science and Engineering, Southeast University, Nanjing, China ({yikun, cheny}@seu.edu.cn).
Y. Zhang is with School of Cybersecurity, Sichuan University, Chengdu, China (yzhang@scu.edu.cn).
Q. Liu is with School of Information Engineering, Nanchang University, Nanchang, China (liuqiegen@ncu.edu.cn).

grounded noise compensation. This limitation restricts the ability to leverage cross-view or projection data for more accurate noise correction, confining improvements to image-domain processing [18]. As a result, globally structured artifacts such as streaking cannot be explicitly addressed, and fine structural details degraded at earlier stages cannot be reliably recovered, which collectively imposes an inherent upper bound on achievable reconstruction fidelity.

To alleviate these limitations, research has moved earlier in the reconstruction pipeline to the post-log domain [8], where measurements are represented as logarithmically transformed line integrals of attenuation. Denoising performed at this stage allows noise and artifacts to be mitigated prior to image reconstruction, thereby offering an effective correction pathway than image-domain processing [19]. Studies include the three-dimensional convolutional network by Yin *et al.* [20], the structure-adaptive filtering approach of Balda *et al.* [21], and the noise-model-aware bilateral filtering method by Manduca *et al.* [22]. Recognizing the complementary strengths of projection-level correction and image-level refinement, a number of recent studies have further explored hybrid reconstruction strategies that jointly leverage information from the post-log projection domain and the image domain, aiming to balance physical consistency with perceptual enhancement. Prominent efforts involve the hybrid-domain integrative transformer iterative network proposed by Wang *et al.* [23], the dual domain closed-loop learning framework introduced by Guo *et al.* [24], and the dual-domain dual-way estimated network developed by Wang *et al.* [25]. Despite these advantages, post-log domain processing exhibits increased sensitivity to restoration accuracy. The logarithmic transformation alters the underlying noise statistics, and residual noises in the restored sinogram are propagated and amplified during filtered back-projection. Such propagated noises often manifest as spatially widespread degradation in the reconstructed image, necessitating substantially higher precision in post-log restoration to preserve overall image quality.

Motivated by the pursuit of physically faithful noise modeling, recent investigations have advanced further upstream to the pre-log projection domain, which operates directly on raw photon counts prior to logarithmic transformation. This domain preserves the original noise statistics and therefore provides the most informative signal representation for principled correction, as demonstrated in recent diffusion-based modeling efforts [26] and tissue-specific Bayesian reconstruction frameworks leveraging texture-dose relationships [27]. Despite these theoretical advantages, reconstruction methods operating in the pre-log domain remain relatively scarce due to several fundamental challenges.

Pre-log sinograms exhibit unbalanced numerical structure, in which background regions correspond to large photon counts while object-related attenuation regions have smaller values. The extreme dynamic range concentrates signal energy in the background, leaving fine structural features as minor variations. This imbalance creates nonuniform gradient landscape for neural networks, in which network parameters tend to be dominated by the high-value background, making learning of subtle structures difficult. Even small deviations in pre-log reconstruction, though negligible relative to the background, are significantly amplified by the logarithmic transformation and subsequent image reconstruction. In particular, regions adjacent to object boundaries that should be close to zero in the post-log domain may acquire positive values, producing a low-level bias across the reconstructed image and generating the "grayish" appearance.

To address these challenges, we propose a novel framework for Pre-Log vOronoi decomposiTion-assisted CT generation, coined PLOT-CT. It introduces a feature disentanglement strategy based on Voronoi decomposition. Pre-log sinogram intensities are partitioned into latent regions corresponding to the high-intensity, low-intensity, and background areas. Within each latent subspace, numerical scales are balanced, gradients are stable, and learning objectives are focused on the features. The latent representations are integrated into a transformer-enhanced diffusion process, which preserves global consistency while maintaining structural fidelity. This design effectively suppresses projection-domain aliasing, mitigates the low-level bias arising from pre-log value disparities, and enables high-fidelity restoration of pre-log CT sinograms.

The theoretical and practical contributions of this work can be summarized as follows:

- A Voronoi-based decomposition mechanism partitions pre-log sinograms into multiple data clusters and constructs corresponding latent spaces, explicitly capturing heterogeneous numerical distributions and addressing the extreme dynamic range in pre-log measurements.
- A diffusion process is performed within the obtained latent spaces, and the resulting latent vectors are used to condition a Transformer for feature reconstruction and refinement. This approach effectively suppresses information aliasing, enhances feature discriminability and structural integrity, and ultimately leads to significantly improved structural preservation and detail reconstruction for pre-log CT sinograms.

The remaining sections of this study are organized as follows: Section II details the motivation and the proposed PLOT-CT framework. The experimental results and comprehensive analysis are shown in Section III. The discussion, including mathematical proofs related to Voronoi space, is provided in Section IV. Finally, we draw conclusions in Section V.

## II. PROPOSED METHOD

### A. Motivation

In CT imaging, the LDCT reconstruction pipeline involves three interconnected data domains, pre-log, post-log, and image domains, each exhibiting distinct gradient behaviors, learning challenges, and noise propagation characteristics.

The pre-log domain, operating directly on raw photon intensity measurements, adheres to the Beer–Lambert law [28], expressed as:

$$I(\theta, s) = I_0 \exp(-P(\theta, s)) + \eta, \quad (1)$$

$$P(\theta, s) = \int \mu(x, y) \mathrm{d}l, \quad (2)$$

where $I(\theta, s)$ denotes the measured photon intensity at projection angle $\theta$ and detector position $s$, $I_0$ is the incident photon intensity, $P(\theta, s)$ represents the noise-free line integral of tissue attenuation, $\mu(x, y)$ is the attenuation coefficient of the tissue at spatial coordinate, $\mathrm{d}l$ is the line element along the projection ray, and $\eta$ is the Poisson measurement noise inherent to photon counting.

Consider the first-order difference between adjacent detector bins:

$$\Delta I = I(s + \Delta s) - I(s). \quad (3)$$

Based on the exponential model, it can be obtained that:
$$\Delta I \approx -I_0 e^{-P} \Delta P, \quad (4)$$
where $\Delta P$ is the difference in true line integrals between adjacent positions. This relationship highlights that the same physical attenuation change $\Delta P$ translates to drastically different gradient magnitudes across varying $P$ intervals, as the gradient scale is dominated by the exponential factor $e^{-P}$.

The absence of a uniform gradient statistical scale in the pre-log domain results in significant scale inconsistency and statistical non-stationarity, making it the most challenging domain for deep learning models to establish stable mappings.

The post-log domain is derived via logarithmic transformation of pre-log data:
$$p(\theta, s) = -\log(I(\theta, s)/I_0) = P(\theta, s) + \varepsilon, \quad (5)$$
where $p(\theta, s)$ is the post-log projection data and $\varepsilon$ is the equivalent additive Gaussian noise converted from the original Poisson noise.

Its adjacent difference is defined as:
$$\Delta p = p(s + \Delta s) - p(s) = \Delta P. \quad (6)$$

Gradient magnitudes directly reflect physical attenuation changes without dependence on the absolute attenuation level $P$, resulting in uniform gradient statistics across the projection space. This linearization of gradients significantly reduces learning difficulty, though global projection coupling remains a challenge.

Image domain is the final stage of the reconstruction pipeline, with image reconstruction expressed as:
$$\hat{\mu} = \mathcal{R}^{-1} p, \quad (7)$$
where $\hat{\mu}$ is the reconstructed attenuation coefficient image and $\mathcal{R}^{-1}$ denotes the Radon inverse transform.

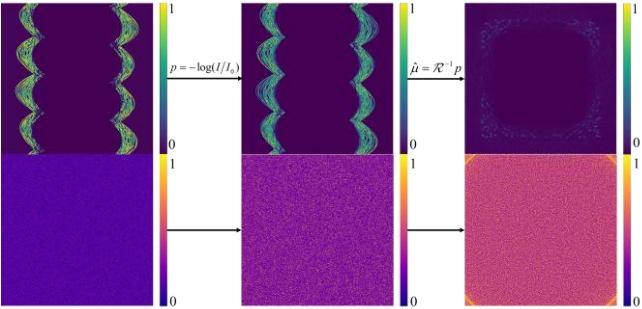

Fig. 2. Visualization of gradients and noise distributions across the pre-log, post-log, and image domain. The first and second rows respectively present the gradient distribution heatmaps and noise distribution heatmaps of the simulated data, and processed in the pre-log domain, post-log domain and image domain.

In image domain, global projection integrals are inverted into local spatial structures, with gradients $\nabla \hat{\mu}(x, y)$ primarily concentrated at true anatomical boundaries. The sparse and spatially stable gradient distribution satisfies deep learning's assumptions of locality and stationarity, making the image domain the easiest to learn among the three. The gradient distribution is shown in the first row of Fig. 2.

The challenges of pre-log domain learning extend beyond gradient inconsistency to a cascading chain of noise amplification. Due to the unstable gradient landscape, pre-log domain modeling inevitably introduces systematic, scale-dependent noises:
$$\delta I = \hat{I} - I, \quad (8)$$
where $\delta I$ is the difference between the model-predicted pre-log intensity $\hat{I}$ and the true intensity $I$.

These noises are not random but correlate with the attenuation level $P$, and their impact is exacerbated by the logarithmic transformation to the post-log domain. Using the post-log definition:
$$p = -\log(I/I_0). \quad (9)$$

The propagation of pre-log domain noises to the post-log domain is approximated as:
$$\delta p = -\log((I + \delta I)/I) \approx -\delta I/I = -\delta I/I_0 e^P, \quad (10)$$
where $\delta p$ is the post-log domain noise. This shows that minor pre-log domain noises are amplified exponentially by the factor $e^P$, particularly for high-attenuation paths. Subsequent image reconstruction propagates these post-log domain noises globally via the Radon inverse transform:
$$\delta \mu = \mathcal{R}^{-1} \delta p. \quad (11)$$
where $\delta \mu$ is the final image domain noise. The globality of the inverse transform ensures that local post-log domain noises spread along ray directions to the entire image, forming a complete stepwise noise-amplified chain.

To intuitively characterize the aforementioned complete noise chain, the second rows of Fig. 2 clearly present the noise plots and Fig. 1 presents corresponding precise probability density distribution plots of the pre-log sinogram. This result explicitly reveals the distinct asymptotic amplification of noises and the distribution pattern followed by the corresponding probability density, thus fully verifying our conclusions.

A fundamental solution to these challenges lies in intensity or attenuation decomposition, leveraging the strict monotonic relationship between pre-log intensity and attenuation $I \leftrightarrow P$.

Using K-means clustering algorithm [29] in Voronoi space to partition pre-log projections into background, low-intensity, and high-intensity regions is mathematically equivalent to segmenting $P$ into distinct intervals. Within each subinterval $k$, the gradient approximation becomes:
$$\Delta I \approx -I_0 e^{-P_k} \Delta P, \quad (12)$$
where $P_k$ is the average true line integral within the interval. This local fixing of the exponential factor stabilizes gradient scales, reduces pre-log domain learning noises $\delta I_k$, and constrains noise amplification in logarithmic transformation, enabling accurate pre-log domain modeling.

Learning in the pre-log domain of CT is fundamentally challenging due to inherent exponential gradient scaling and complex statistical inhomogeneities. This difficulty gives rise to significant scale-dependent noises, which are amplified by the logarithmic transformation and propagated globally during standard image reconstruction. Intensity-based decomposition of the pre-log domain effectively stabilizes gradient statistics, enabling accurate pre-log domain modeling.

### B. Overview of PLOT-CT

PLOT-CT comprises core components: Pre-log sinogram decomposition block with Voronoi space-based, multi-latent space representation via diffusion block, multi-latent space processing via Transformer block, and overall reconstruction. A spatial decomposition strategy partitions pre-log sinograms into feature clusters through K-means clustering. It employs the statistical distribution of pixel intensity in pre-log sinograms to realize precise partition and build Voronoi space in the latent space. These clusters form independent regions in latent Voronoi space to separate feature representations. Voronoi regions possess distinct boundaries defined by neighborhood relationships to ensure high feature consistency within each region. This decomposition covers high-intensity, low-intensity and background regions with statistical properties including different intensity variance ranges and pixel distribution patterns. It lays foundation for subsequent targeted feature processing. The physical phenomena corresponding to the sub-regions of the pre-log sinogram after decomposition in Voronoi space are shown in Fig. 3. This figure shows the distribution of gradients and errors in different Voronoi regions after decomposition. Gradients vary drastically only at region junctions and noises distribute more uniformly within each region after partition. This phenomenon enables the model to perform targeted learning on each component.

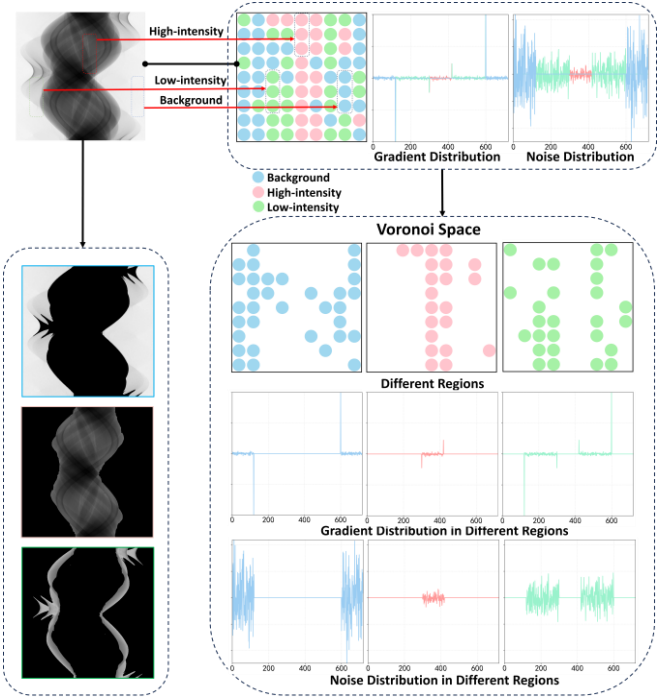

Fig. 3. The relationship of pre-log sinograms and information clusters, the distributions followed by data gradients and noise, as well as the gradient and noise distributions of information clusters and their corresponding data after decomposition in Voronoi space.

In the training phase, normal-dose and low-dose pre-log sinograms first undergo Voronoi decomposition via K-means clustering, forming three feature clusters. Next, Transformer block is pre-trained, taking low-dose cluster features as inputs and normal-dose cluster features as supervision targets to learn cross-cluster feature mapping. Finally, diffusion model is trained following a Markov chain [30] with forward noise addition and reverse denoising to suppress noise and establish structural priors for reconstruction.

In the inference phase of LDCT image processing, the input low-dose pre-log sinogram contains noise from reduced radiation and first decomposes into three Voronoi clusters via the pre-log sinogram decomposition block (PSDB) to isolate data characteristics for targeted processing. These three clusters are concatenated into a multi-channel input to enrich the information dimension. The diffusion block performs denoising following Markov chain reverse denoising, gradually suppressing noise while retaining key structural details. The Transformer block uses multi-head transposed attention (MTA) to capture long-range dependencies across Voronoi space and a gated feed-forward network (GFN) to enhance feature discriminability, refining cluster features. Eventually, the three refined features are fused via a combination strategy, and the pre-log sinogram is converted to the post-log domain via logarithmic transformation for high-quality CT image generation. Then the filtered back projection (FBP) [31] algorithm is applied to generate the high-quality CT image.

### C. Training Procedure

The pipeline of the PLOT-CT training procedure is illustrated in Fig. 4. This comprehensive procedure integrates Voronoi decomposition, Transformer block pre-training, and Transformer-based diffusion modeling into a unified framework for LDCT reconstruction, with explicit optimization objectives for each stage to enhance feature disentanglement and denoising performance.

Initially, both normal-dose pre-log sinogram $Y_{ND}$ and low-dose pre-log sinogram $Y_{LD}$ undergo Voronoi decomposition through K-means clustering to extract distinct feature representations. Each sinogram is partitioned into three characteristic regions corresponding to high-intensity, low-intensity, and background areas. This decomposition process can be mathematically represented as:

$$Y_{ND} = \bigcup_{k=1}^{3} C_{ND,k}, \quad C_{ND,k} = \text{K-means}(Y_{ND}, k), \quad (13)$$

$$Y_{LD} = \bigcup_{k=1}^{3} C_{LD,k}, \quad C_{LD,k} = \text{K-means}(Y_{LD}, k), \quad (14)$$

where $C_{ND,k}$ and $C_{LD,k}$ denote the $k$-th feature cluster ($k=1$ for high-intensity, $k=2$ for low-intensity, $k=3$ for background) of normal-dose and low-dose sinograms, respectively.

In the pre-training stage, the Transformer block architecture undergoes supervised training using $C_{LD,k}$ and $C_{ND,k}$ to learn cross-cluster feature mapping. The decomposed features from both $Y_{LD}$ and $Y_{ND}$ are structured into multi-channel inputs to retain cluster-specific features, i.e.,

$$F_{l,k} = C_{LD,k}, \quad F_{target,k} = C_{ND,k}, \quad (15)$$

$$F_l = Concat(F_{l,1}, F_{l,2}, F_{l,3}), \quad (16)$$

$$F_{target} = Concat(F_{target,1}, F_{target,2}, F_{target,3}), \quad (17)$$

$$F_{in} = Concat(F_l, F_{target}), \quad (18)$$

where $Concat$ denotes channel-wise concatenation. $F_{l,k}$ directly corresponds to the $k$-th feature cluster of low-dose sinograms obtained by K-means decomposition, and $F_{target,k}$ is the $k$-th feature cluster of normal-dose sinograms. The model takes $F_{in}$ as input and $F_{target}$ as the ground truth, learning the cross-cluster feature mapping from low-dose to normal-dose.

Following channel-wise concatenation, a three-channel latent vector in the Voronoi space is obtained through the latent encoder:

$$V_{in,k} = \text{Latent Encoder}(F_{in,k}), \quad (19)$$

Subsequently, $V_{in,k}$ is fed into the MTA and GFN to guide image reconstruction:

$$\hat{M}_{in,k} = W_l^1 V_{in,k} \odot \text{Norm}(M) + W_l^2 V_{in,k}, \quad (20)$$

where $\hat{M}_{in,k}$ and $M$ denote the input feature map and the output feature map, respectively, Norm represents the normalization layer, $W_l$ refers to the linear layer, and $\odot$ stands for element-wise multiplication.

The Transformer block is trained in a supervised manner to learn the cross-cluster feature mapping relationship. The core components of the Transformer block include MTA for capturing long-range dependencies between different regions in Voronoi space and GFN for enhancing discriminative features.

In MTA module, the input feature $M_{in,k}$ is first projected into three components: Query $Q = W_d^Q W_c^Q M_{in,k}$, Key $K = W_d^K W_c^K M_{in,k}$, and Value $V = W_d^V W_c^V M_{in,k}$, where $Q, K, V$ are generated via a combination of 1×1 point-wise convolution ($W_c$) and 3×3 depth-wise convolution ($W_d$) to capture both channel-wise and spatial information.

Once $Q, K, V$ are generated, they are split into $H$ attention heads to capture multi-scale dependencies. For each head $h$, the scaled dot-product attention is computed, and the results of all heads are concatenated to obtain the final MTA output, as shown:

$$\text{MTA}(M_{in,k}) = Concat(\text{Softmax}(Q_h K_h / \sqrt{d_k})V_h)_{h=1}^H, \quad (21)$$

where $Q_h, K_h, V_h$ denote the $Q, K, V$ of the $h$-th attention head, $d_k$ represents the dimension of each attention head to avoid the gradient vanishing issue caused by large dot-product values, and the dot product of $Q_h$ and $K_h$ is scaled by $\sqrt{d_k}$. $H$ represents the total number of attention heads, and Softmax is used to normalize the attention weights, ensuring the sum of weights for each position is 1.

GFN module enhances the discriminative ability of features through non-linear transformations, combining 1×1 point-wise convolution for channel interaction and 3×3 depth-wise convolution for local spatial interaction with a gating mechanism, as shown:

$$\text{GFN}(M) = \text{GELU}(W_d^1 W_c^1 M) \odot W_d^2 W_c^2 M + M, \quad (22)$$

where $M$ is the output of MTA module, $W_c^1, W_c^2$ denote 1×1 point-wise convolution kernels, $W_d^1, W_d^2$ denote 3×3 depth-wise convolution kernels, and GELU, the Gaussian Noise Linear Unit, is the activation function to introduce non-linearity.

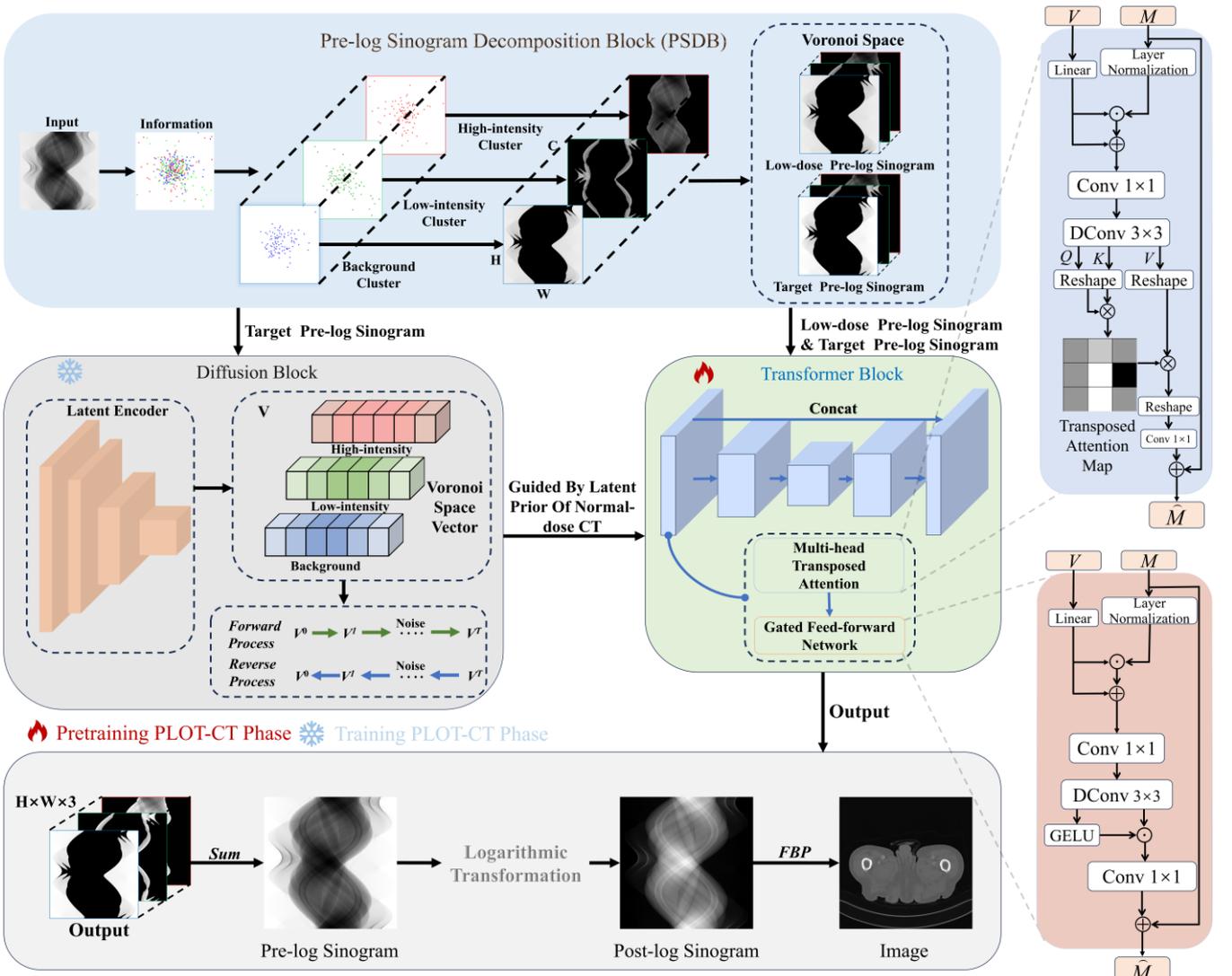

Fig. 4. The training and inference pipeline of PLOT-CT. PLOT-CT mainly consists of Pre-log Sinogram Decomposition Block (PSDB), Diffusion Block and Transformer block. Pre-log sinogram is decomposed and processed to construct features, which are then used to guide the Transformer block Structure to reconstruct the final result. The decomposed sinogram features are processed through diffusion-based denoising and transformer refinement, which are then fused and transformed to reconstruct high-quality CT images.

After completing the pre-training, the diffusion model is trained to denoise $F_{in,k}$, leveraging the structural prior from the Transformer block. The diffusion process follows a Markov chain, where the forward process adds noise to clean $F_{target,k}$, and the reverse process trains the diffusion transformer to recover clean clusters by estimating noise.

In forward diffusion process, over $T$ time steps, noise is gradually added to $V_{in,k}$ to generate noisy $V_{k,t}$ at each step. The forward process follows a Gaussian transition, as shown:

$$q(V_{k,t}|V_{k,t-1}) = \mathcal{N}(V_{k,t}; \sqrt{\alpha_t}V_{k,t-1}, \beta_t \mathbf{I}), \quad (23)$$

where $\alpha_t = 1 - \beta_t$, with $\beta_t$ being the noise scheduling parameter that increases linearly from 0.1 to 0.99 over $T$ steps. Additionally, $\mathcal{N}(\mu, \sigma^2 \mathbf{I})$ denotes a Gaussian distribution with mean $\mu$ and covariance $\sigma^2 \mathbf{I}$, and $\mathbf{I}$ is the identity matrix.

The reverse process uses diffusion transformer to estimate the noise $\epsilon$ in $V_{k,t}$ and recover the cluster at step $t-1$ ($\hat{V}_{k,t-1}$). The diffusion transformer takes $V_{k,t}$, the time step t, and the conditional feature $V_{in,k}$ as inputs to predict noise, as shown:

$$\hat{V}_{k,t-1} = (V_{k,t} - \epsilon_\theta(V_{k,t}, t, V_{in,k})(1-\alpha_t)/\sqrt{1-\alpha_t})/\sqrt{\alpha_t}, \quad (24)$$

where $\epsilon_\theta$ is the noise estimation network with parameters $\theta$, $(1-\alpha_t)/\sqrt{1-\alpha_t}$ is the noise scaling factor and $1/\sqrt{\alpha_t}$ normalizes the mean of the reverse Gaussian distribution.

### D. Inference Procedure

The inference pipeline of PLOT-CT is visually illustrated in Fig. 4, which systematically integrates key sinogram decomposition, multi-channel feature construction, diffusion transformer denoising, feature fusion, domain transformation, and FBP to reconstruct refined high-quality CT images from raw low-dose pre-log sinograms.

First, the input $Y_{LD}$ is fed into the PSDB. Following the K-means clustering strategy defined in the training procedure, $Y_{LD}$ is partitioned into three distinct information clusters corresponding to high-intensity, low-intensity, and background regions.

The decomposition process is rephrased for reconstruction as:

$$Y_{LD} = \bigcup_{k=1}^{3} C_{LD,k}, \quad C_{LD,k} = \text{PSDB}(Y_{LD}, k), \quad (25)$$

$$F_{low,k}^{\text{pre-log}} = C_{LD,k}, \quad (26)$$

where in PLOT-CT with $n=3$, $k=1,2,3$ denotes high-intensity, low-intensity, and background clusters, respectively. $F_{low,k}^{\text{pre-log}}$ is low-dose pre-log sinogram of the $k$-th Voronoi space.

To fully leverage the structural information of each region, the cluster-specific features $F_{low,k}^{\text{pre-log}}$ are concatenated along the channel dimension to form a three-channel input feature map for PLOT-CT model. This design ensures that the model simultaneously processes distinct intensity clusters and captures cross-cluster dependencies.

The three-channel input $F_{in}$ is mathematically defined as:

$$F_{in} = Concat(F_{low,1}^{\text{pre-log}}, F_{low,2}^{\text{pre-log}}, F_{low,3}^{\text{pre-log}}). \quad (27)$$

The $F_{in}$ is fed into the PLOT-CT model, which consists of three core modules including PSDB, the diffusion block, and the Transformer block.

Following the reverse denoising process of the Markov chain, the diffusion block estimates the noise in each cluster's features using the Transformer-based network $\epsilon_\theta$. For the $i$-th cluster, the noise estimation process is:

$$\hat{\epsilon}_t^i = \epsilon_\theta(F_t^i, t, F_{in}^i), \quad (28)$$

where $F_t^i$ is the noisy feature of the $i$-th cluster at time step $t$, and $F_{in}^i$ is the $i$-th channel of $F_{in}$. The diffusion block iteratively removes noise to generate denoised cluster features $\hat{C}^i$.

Organized in a U-Net architecture, this block uses MTA from Eq. (21) to capture long-range dependencies across Voronoi space, and GFN as outlined in Eq. (22) to enhance discriminative features. It further refines $\hat{C}^i$ to ensure structural integrity.

Since the three reconstructed cluster features $\hat{C}^1$, $\hat{C}^2$, $\hat{C}^3$ are non-overlapping and each corresponds to a distinct region, they are fused via direct element-wise summation to reconstruct a complete pre-log sinogram $\hat{Y}^{\text{pre-log}}$.

The fusion process is defined as:

$$\hat{Y}^{\text{pre-log}} = \hat{C}^1 + \hat{C}^2 + \hat{C}^3. \quad (29)$$

Traditional CT reconstruction algorithms typically require input data in post-log domain. Therefore, reconstructed $\hat{Y}^{\text{pre-log}}$ is converted to post-log domain via a logarithmic transformation.

The transformation process is defined as:

$$\hat{Y}^{\text{post-log}} = -\ln(\hat{Y}^{\text{pre-log}}/I_0), \quad (30)$$

where $I_0$ denotes the photon intensity, and $\hat{Y}^{\text{post-log}}$ is the reconstructed post-log sinogram. This step converts the multiplicative Poisson noise in pre-log domain to approximately additive Gaussian noise in post-log domain.

The final step uses FBP to convert $\hat{Y}^{\text{post-log}}$ from post-log domain to image domain, generating a high-quality CT image. The FBP process in PLOT-CT is mathematically summarized as a single formulation, which integrates core operations such as filtering and back-projection while aligning with CT imaging physics:

$$\mu_{final} = \text{FBP}(\hat{Y}^{\text{post-log}}, H_{filter}, \Omega), \quad (31)$$

where $\mu_{final}$ is the final reconstructed CT image, $H_{filter}$ denotes the filter used in FBP and $\Omega$ represents the set of projection angle.

---

**Algorithm 1: PLOT-CT**

**Require**: Low-dose pre-log sinogram $Y_{LD}$

1: **Initialization**: Set diffusion steps $T$, noise schedule $\alpha_t$, $\bar{\alpha}_t$. Decompose $Y_{LD}$ into 3 Voronoi clusters. Construct 3-channel input $F = [C_{high}, C_{low}, C_{bg}]$.

2: $V = \text{Latent Encoder}(F)$

3: **For** $t = T$ to 1 do

4:     **For** each cluster $i$:

5:         **Estimate noise**: $\epsilon_t^i = \epsilon_\theta(V_t^i, t, \text{cond}^i)$

6:         **Update**: $V_{i-1}^i = (V_i^i - (1-\alpha_t)/\sqrt{1-\alpha_t}\epsilon_t^i)/\sqrt{\alpha_t}$

7:     **End For**

8: **End For**

9: **Refine clusters**: $\hat{C}^i = \text{TransformerBlock}(V_0^i)$

10: **Fuse**: $\hat{Y}^{\text{pre-log}} = \sum_i \hat{C}^i$

11: **Convert**: $\hat{Y}^{\text{post-log}} = -\ln(\hat{Y}^{\text{pre-log}}/I_0)$

12: **Reconstruct**: $\mu_{final} = \text{FBP}(\hat{Y}^{\text{post-log}}, H_{filter}, \Omega)$

13: **Return** $\mu_{final}$

## III. EXPERIMENTS

### A. Data Specification

*AAPM Challenge Dataset:* This dataset was provided by Mayo Clinic for the AAPM LDCT Grand Challenge and includes simulated human abdominal CT images for evaluating relevant low-dose imaging algorithms [32]. It comprises 4742 normal-dose CT images from 10 clinical patients. For fan-beam CT reconstruction, Siddon's ray-driven algorithm [33] was used, with a rotation center 40 cm from both source and detector, a precise 41.3 cm-wide detector consisting of 720 elements, and over 360 uniformly distributed projection views. The ground truth was reconstructed from normal-dose projections using FBP. Among then, 2200 images were used for training, with 5% of them designated as the dedicated validation set, while 10% of the remaining 2542 images were employed for formal testing.

*CHAOS CT Dataset:* For the generalization study, we selected 15% of the total images from the Combined Healthy Abdominal Organ Segmentation (CHAOS) dataset, a well-recognized classic benchmark for clinical abdominal imaging. This dataset consists of 40 high-quality clinical CT scans of healthy subjects, acquired with a 1 mm slice thickness and a 512×512 resolution under standardized abdominal contrast protocols.

### B. Experimental Settings

PLOT-CT model was implemented with ODL and PyTorch [34] on a workstation equipped with two NVIDIA RTX 3090 GPUs, which ensures computational power and stability for model training and inference. To simulate X-ray tomography, Poisson noise from photon counting statistics was introduced to projection data, and X-ray sources with varying intensity levels were configured to match different clinical low-dose acquisition conditions. Two-stage training strategy balanced convergence speed and performance: the first stage used Adam with an initial 1.5e-4 learning rate, paired with CosineAnnealingRestartLR to prevent premature convergence. The second stage used Adam with an initial 2e-4 learning rate, with a Multi-StepLR scheduler that halves the rate at 300k steps to refine parameters. For evaluation of reconstruction quality, three standard LDCT metrics Peak Signal-to-Noise Ratio (PSNR), Structural Similarity Index (SSIM), and Mean Squared Noise (MSE) were used. Values were averaged over test samples for reliability. Per conventional standards, higher PSNR/SSIM and lower MSE indicate better quality. The source code of PLOT-CT and experimental results are available for reproduction at: https://github.com/yqx7150/PLOT-CT.

### C. Experimental Evaluation

*AAPM Challenge Dataset:* We compare PLOT-CT with six baseline techniques in LDCT reconstruction, including Pix2Pix [35], NCSN++ [36], CaGAN [37], Re-UNet [38], WiTUnet [39] and SDCNN [40] to assess their performance. We selected 2200 images from the AAPM challenge dataset, a used benchmark for LDCT research, to ensure the experiment's validity and comparability, and added three noise levels corresponding to 1e4, 5e3, and 1e3 photons along each X-ray path to simulate different low-dose scenarios. After adding noise, 6600 images were obtained, and PLOT-CT was trained using these pre-log sinogram data. The hyperparameters of all baselines were set following guidelines from original studies to guarantee a fair comparative environment. For experiments, we configured the X-ray system to emit 1e4, 5e3, and 1e3 photons along each path to replicate LDCT acquisition conditions, and Table I presents quantitative evaluation results including PSNR, SSIM, and MSE of reconstructed images, with the best values highlighted in bold. Notably, images generated by PLOT-CT exhibit fewer artifacts and lower noise than baselines, helping maintain anatomical structures critical for clinical analysis.

TABLE I
RECONSTRUCTION PSNR/SSIM/MSE OF AAPM CHALLENGE DATA USING DIFFERENT METHODS AT DIFFERENT NOISE LEVELS.

| Method | $a_i = $ 1e4 | | | $a_i = $ 5e3 | | | $a_i = $ 1e3 | | |
|---|---|---|---|---|---|---|---|---|---|
| | PSNR ↑ | SSIM ↑ | MSE ↓ | PSNR ↑ | SSIM ↑ | MSE ↓ | PSNR ↑ | SSIM ↑ | MSE ↓ |
| FBP [31] | 27.77 | 0.6666 | 1.73e-3 | 25.24 | 0.6166 | 3.11e-3 | 19.64 | 0.5164 | 1.11e-2 |
| Pix2Pix [35] | 33.28 | 0.9238 | 1.02e-3 | 31.10 | 0.9037 | 1.51e-3 | 26.18 | 0.8013 | 4.44e-3 |
| NCSN++ [36] | 37.45 | 0.9100 | 2.94e-4 | 35.91 | 0.8691 | 4.26e-4 | 33.89 | 0.8277 | **1.00e-3** |
| CaGAN [37] | 32.41 | 0.9133 | 7.92e-4 | 32.32 | 0.9033 | 8.70e-4 | 31.06 | 0.8498 | 1.18e-3 |
| Re-UNet [38] | 31.84 | 0.9361 | 1.69e-3 | 30.91 | 0.9285 | 1.96e-3 | 28.74 | 0.8937 | 3.04e-3 |
| WiTUnet [39] | 39.83 | 0.9702 | 2.21e-4 | 37.38 | 0.9581 | 4.12e-4 | 31.28 | 0.9249 | 1.91e-3 |
| SDCNN [40] | 32.09 | 0.9325 | 1.46e-3 | 30.84 | 0.9106 | 1.78e-3 | 29.11 | 0.8007 | 2.21e-3 |
| **PLOT-CT** | **42.19** | **0.9721** | **6.38e-5** | **40.44** | **0.9664** | **1.42e-4** | **34.62** | **0.9429** | 1.14e-3 |

To provide a comprehensive evaluation of our proposed approach, this study conducts both visual and quantitative assessments of the algorithms' performance. Table I presents detailed evaluation results recording PSNR, SSIM, and MSE under different noise levels, with PLOT-CT achieving the highest PSNR and SSIM and the lowest MSE across all scenarios. Specifically, Figs. 5 to 7 display the reconstructed images and their corresponding residual maps under multiple noise levels. These residual images reflect the reconstruction noise distribution to capture performance differences when processing complex textures or edge regions. This format enables a direct visual comparison of noise suppression and detail preservation capabilities, allowing researchers to observe how each method balances noise reduction and feature retention without relying solely on quantitative metrics. Such visual insights, combined with the quantitative superiority shown in Table I, demonstrate the advantages of the experimental results and offer a reference for future algorithm optimization.

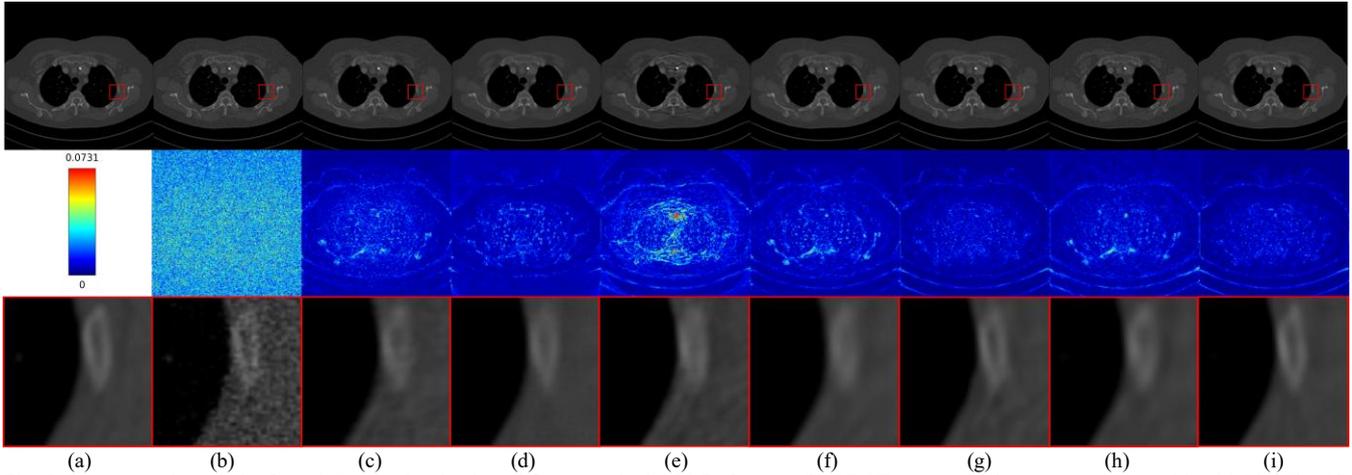

Fig. 5. Reconstruction results from 1e4 noise level using different methods. (a) Reference, (b) LDCT, versus the images reconstructed by (c) Pix2Pix, (d) NCSN++, (e) CaGAN, (f) Re-UNet, (g) WiTUnet, (h) SDCNN, (i) PLOT-CT. The display windows are [-85, 50] HU. The second row shows the residual maps of the reconstruction, and the third row shows the enlarged view of the ROI (indicated by the red box in the first row).

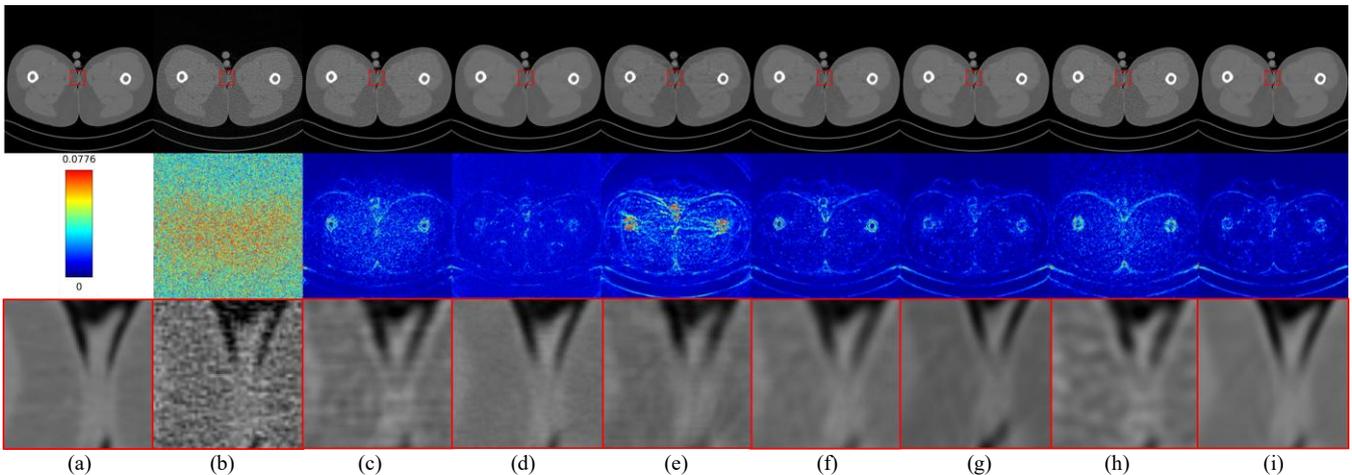

Fig. 6. Reconstruction results from 5e3 noise level using different methods. (a) Reference, (b) LDCT, versus the images reconstructed by (c) Pix2Pix, (d) NCSN++, (e) CaGAN, (f) Re-UNet, (g) WiTUnet, (h) SDCNN, (i) PLOT-CT. The display windows are [-185, 105] HU. The second row shows the residual maps of the reconstruction, and the third row shows the enlarged view of the ROI.

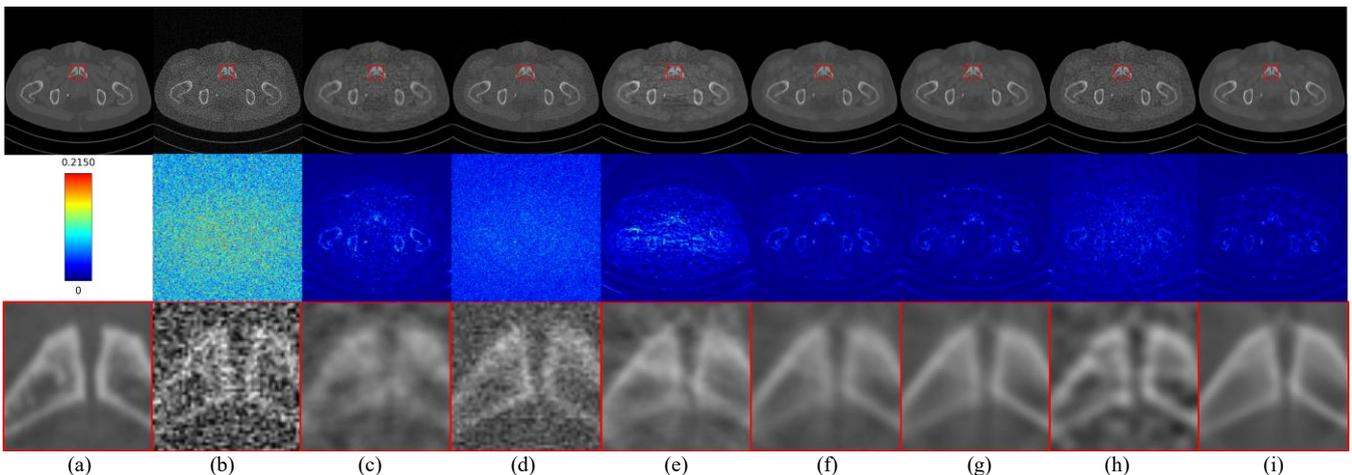

Fig. 7. Reconstruction results from 1e3 noise level using different methods. (a) Reference, (b) LDCT, versus the images reconstructed by (c) Pix2Pix, (d) NCSN++, (e) CaGAN, (f) Re-UNet, (g) WiTUnet, (h) SDCNN, (i) PLOT-CT. The display windows are [-105, 75] HU. The second row shows the residual maps of the reconstruction, and the third row shows the enlarged view of the ROI.

Among all the comparative algorithms tested in our study, Pix2Pix shows noticeable structural flaws in its generated results even under relatively low noise levels. NCSN++ often over-smooths edges during denoising and at extremely high noise levels it can barely remove noise effectively and frequently introduces stripe artifacts in high-contrast regions. SDCNN achieves only moderate noise reduction and cannot fully recover fine textures and may leave subtle artifacts in the reconstructed images. Re-UNet maintains basic structural integrity under moderate noise but still fails to adequately capture texture details. CaGAN struggles to restore complex structural details and cannot accurately reproduce intricate textures and hierarchical structures. WiTUnet delivers strong performance ranking second only to PLOT-CT while still exhibiting some noise sensitivity which results in minor artifacts and incomplete structures in reconstructions and prevents accurate recovery of fine image details. Across all tested noise levels and scenarios, PLOT-CT consistently outperformed comparative methods in both quantitative metrics and visual quality, demonstrating robust stability.

*CHAOS Dataset:* To better explore the generalization of PLOT-CT, we apply the comparative methods to the CHAOS dataset a widely used medical imaging benchmark that covers diverse clinical abdominal imaging scenarios to ensure comprehensive evaluation. Furthermore, as an unsupervised model, NCSN++ is trained on 2200 images from the AAPM challenge dataset while all other models in comparison are trained under noise levels of 1e4, 5e3, and 1e3 to simulate realistic LDCT acquisition conditions. Detailed quantitative evaluations of PSNR, SSIM, and MSE metrics are presented in Table II. PLOT-CT outperforms all comparative methods across key performance indicators, demonstrating robust generalization and superior performance. Additionally, Fig. 8 displays representative reconstruction results and noise maps of PLOT-CT, SDCNN, and other comparative methods to visually verify the effectiveness of the proposed approach.

TABLE II
RECONSTRUCTION PSNR/SSIM/MSE OF CHAOS CT DATA USING DIFFERENT METHODS AT DIFFERENT NOISE LEVELS.

| Method | $a_i = 1e4$ | | | $a_i = 5e3$ | | | $a_i = 1e3$ | | |
|---|---|---|---|---|---|---|---|---|---|
| | PSNR ↑ | SSIM ↑ | MSE ↓ | PSNR ↑ | SSIM ↑ | MSE ↓ | PSNR ↑ | SSIM ↑ | MSE ↓ |
| FBP [31] | 31.20 | 0.7472 | 7.89e-4 | 28.51 | 0.6837 | 1.45e-3 | 22.24 | 0.5722 | 6.15e-3 |
| Pix2Pix [35] | 33.02 | 0.8491 | 5.19e-4 | 32.18 | 0.8210 | 6.28e-4 | 28.94 | 0.6682 | 1.31e-3 |
| NCSN++ [36] | 34.06 | 0.8850 | 4.43e-4 | 33.52 | 0.8618 | 4.98e-4 | 29.43 | 0.6207 | 1.18e-3 |
| CaGAN [37] | 31.19 | 0.8884 | 8.68e-4 | 30.54 | 0.8827 | 1.00e-3 | 29.10 | 0.8533 | 1.47e-3 |
| Re-UNet [38] | 33.49 | 0.9329 | 4.91e-4 | 33.27 | 0.9262 | 5.18e-4 | 31.25 | 0.8950 | 9.27e-4 |
| WiTUnet [39] | 38.50 | 0.9583 | 1.47e-4 | 37.03 | 0.9504 | 2.29e-4 | 33.64 | 0.9274 | 4.98e-4 |
| SDCNN [40] | 35.75 | 0.9337 | 2.81e-4 | 34.80 | 0.9100 | 3.14e-4 | 31.11 | 0.7931 | 8.39e-4 |
| **PLOT-CT** | **38.77** | **0.9627** | **1.40e-4** | **37.83** | **0.9555** | **1.74e-4** | **34.05** | **0.9350** | **4.56e-4** |

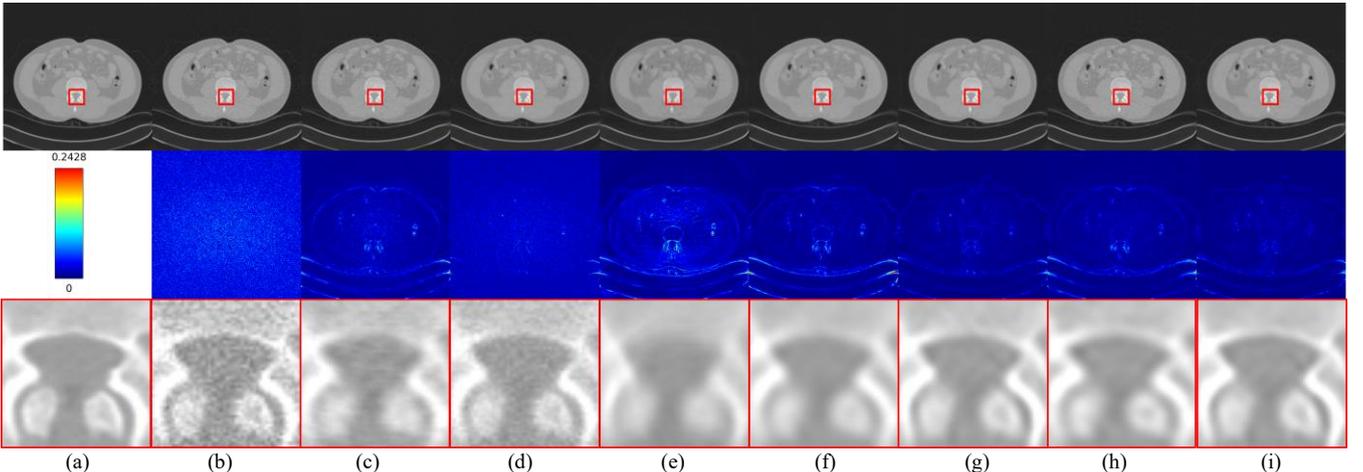

Fig. 8. Reconstruction results from 1e3 noise level using different methods. (a) Reference, (b) LDCT, versus the images reconstructed by (c) Pix2Pix, (d) NCSN++, (e) CaGAN, (f) Re-UNet, (g) WiTUnet, (h) SDCNN, (i) PLOT-CT. The display windows are [-155, 145] HU. The second row shows the residual maps of the reconstruction, and the third row shows the enlarged view of the ROI.

### D. Ablation Study

We conduct a detailed ablation study on multiple key modules of the PLOT-CT and investigate the effectiveness of the proposed clustering method in LDCT reconstruction.

*Different Numbers of Clusters:* In this subsection, we primarily explore the specific impact of different numbers of information cluster partitions on reconstruction performance, as well as how to effectively enhance such performance using the same well designed clustering strategy. Specifically, we conduct comparative investigations on four typical cluster partition scenarios, namely one cluster, two clusters, three clusters and four clusters, which cover the main partition categories concerned in this study. We adopt several common evaluation metrics including PSNR, SSIM and RMSE. PSNR effectively reflects the peak signal-to-noise ratio of reconstructed images, SSIM accurately measures the structural similarity between reconstructed images and reference images, and RMSE reliably quantifies the root mean squared noise between pixel values of the two images. The corresponding experimental results are summarized in Fig. 9.

To ensure rigorous fair comparison and safeguard the comprehensive reliability, repeatability, and comparability of results in this LDCT reconstruction study, we standardized the experimental conditions and unified the uniform detailed operational procedures. Each method was trained on the same training sample and tested using an identical test dataset, eliminating any potential bias from differences in training samples or variations in test data. LDCT reconstruction was performed under a predetermined fixed noise level of 1e4, with clustering serving as the key core baseline method. This noise level was selected based on conventional common LDCT reconstruction parameters and noise control requirements, ensuring the experimental scenario's rationality and representativeness. This systematic thorough approach examines the impact of four-category clustering on reconstruction, providing comprehensive data for subsequent comparisons and analysis of clustering strategies.

As shown in Fig. 9, the relevant method that effectively partitioning the data into three clusters significantly outperforms the other existing partitioning schemes across all critical metrics, including PSNR, SSIM, and MSE, achieving the optimal reconstruction results. This strongly confirms that the triple-cluster partitioning strategy effectively balances the inherent trade-off between feature separation and reconstruction fidelity, making it the optimal preferred approach adopted in our proposed method.

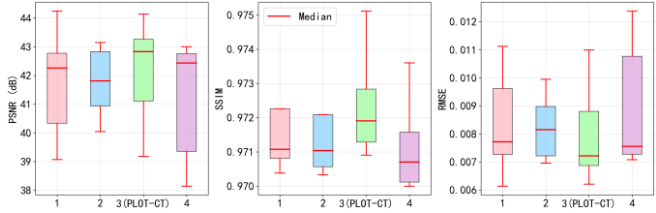

Fig. 9. Comparison of reconstruction PSNR/SSIM/RMSE on AAPM challenge dataset with different numbers of clusters after Voronoi decomposition.

Under noise level 1e4, reconstruction results with different cluster numbers are depicted in Fig. 9. All one two four-cluster partitioning strategies exhibit noise susceptibility. Two-cluster partitioning fails to distinguish tissue intensity levels, causing blurred boundaries between soft tissues and dense structures. Four-cluster approach introduces interference from redundant divisions, increasing local artifacts and compromising structural continuity. One-cluster strategy without partitioning performs poorly in noise suppression. Reconstructed images via this approach contain visible noise and streak artifacts common LDCT issues along with structural information loss obscuring anatomical features. In contrast, the proposed PLOT-CT shown in Fig. 10 demonstrates superior noise reduction, effectively suppressing streak and Gaussian noise while retaining high-fidelity details.

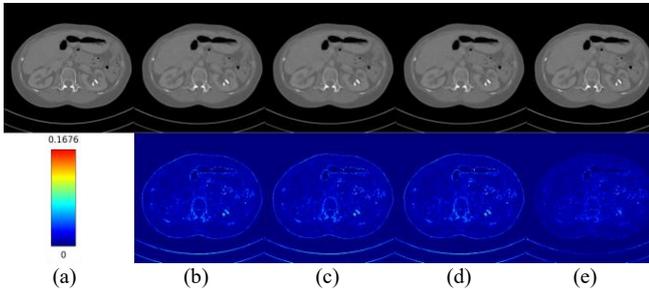

Fig. 10. Reconstruction results at a noise level of 1e4 with different numbers of information clusters. (a) Reference versus the images reconstructed using (b) 1 information cluster, (c) 2 information clusters, (d) 4 information clusters, and (e) 3 information clusters. The display windows are [-90, 70] HU. The second row shows the residual maps of the reconstruction.

***Different Components in PLOT-CT Model:*** This study evaluates the contribution of two components in the PLOT-CT framework via ablation experiments. As demonstrated in Table III, the complete PLOT-CT model consistently achieves optimal performance across all noise levels, 1e4, 5e3, and 5e2 photons. Notably, the model also attains the highest PSNR, SSIM and MSE even under untrained noise levels, and this result verifies capability in handling noise scenarios beyond trained conditions. Removing the diffusion block results in performance degradation, evident under higher noise conditions. Similarly, eliminating the K-means clustering component leads to performance deterioration, which confirms that decomposing projection data into intensity clusters effectively mitigates information aliasing and enhances feature representation. These ablation findings not only corroborate contributions of each component but also underscore rationality of collaborative design in reinforcing adaptability of the PLOT-CT framework to varying noise intensities, while consistent performance gaps between the complete model and ablated variants further validate rigor of the evaluation paradigm.

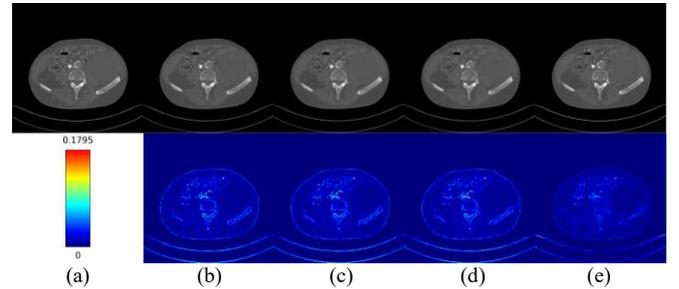

Fig. 11. Reconstruction results from 1e4 noise level using different methods. (a) Reference versus the images reconstructed by (b) (w/o) Diffusion Block & PSDB, (c) (w/o) Diffusion Block, (d) (w/o) PSDB, (e) PLOT-CT. The display windows are [-90, 70] HU. The second row shows the residual maps of the reconstruction.

The visual evidence presented in Fig. 11 provides compelling support for these quantitative findings. The reconstruction results clearly show that the complete PLOT-CT model achieves superior structural preservation with minimal residual noises, especially in regions containing fine anatomical details. In contrast, the ablated versions demonstrate varying degrees of artifact presence and detail loss. This comprehensive analysis validates that integrating K-means driven sinogram decomposition with the transformer-based diffusion process creates a powerful framework for LDCT reconstruction. The clustering strategy provides structurally organized inputs to the diffusion model, which in turn generates high-quality priors. This synergy ultimately enables robust performance across diverse noise conditions.

TABLE III
RECONSTRUCTION PSNR/SSIM/MSE OF AAPM CHALLENGE DATA USING DIFFERENT METHODS AT DIFFERENT NOISE LEVELS.

| Noise level | (w/o) Diffusion Block & PSDB | | | (w/o) Diffusion Block | | | (w/o) PSDB | | | **PLOT-CT** | | |
| --- | --- | --- | --- | --- | --- | --- | --- | --- | --- | --- | --- | --- |
| | PSNR ↑ | SSIM ↑ | MSE ↓ | PSNR ↑ | SSIM ↑ | MSE ↓ | PSNR ↑ | SSIM ↑ | MSE ↓ | PSNR ↑ | SSIM ↑ | MSE ↓ |
| $a_i$ = 1e4 | 41.16 | 0.9714 | 9.37e-5 | 41.67 | 0.9714 | 7.36e-5 | 41.68 | 0.9714 | 7.28e-5 | **42.19** | **0.9721** | **6.38e-5** |
| $a_i$ = 5e3 | 40.09 | 0.9647 | 1.52e-4 | 39.95 | 0.9656 | 1.67e-4 | 39.98 | 0.9647 | 1.73e-4 | **40.44** | **0.9664** | **1.42e-4** |
| $a_i$ = 5e2 | 31.51 | 0.9248 | 2.11e-3 | 31.61 | 0.9285 | 2.14e-3 | 31.45 | 0.9261 | 2.21e-3 | **31.95** | **0.9318** | **1.96e-3** |

## IV. DISCUSSION

This study proposes PLOT-CT for LDCT reconstruction via Voronoi space decomposition in pre-log projection domain. Using K-means to decompose pre-log sinograms into Voronoi space with distinct intensities and constructing multi-latent representations, our approach disentangles features and suppresses aliasing. To provide theoretical justification for the proposed Voronoi decomposition strategy, particularly regarding the mathematical basis for selecting the number of clusters, this section presents a mathematical proof demonstrating the existence of a local optimal cluster number $k^*$ in Voronoi space decomposition that achieves the local balance between reconstruction noise and model complexity.

Let pre-log sinogram data space be a compact set $\mathcal{X} \subset \mathbb{R}^d$, and let $\mu$ be a probability measure on $\mathcal{X}$ describing the data distribution. A Voronoi decomposition with $k$ generators $P=\{p_1, p_2, ..., p_k\} \subset \mathbb{R}^d$ partitions $\mathcal{X}$ into $k$ Voronoi cells $V_1, V_2, \cdots, V_k$, where each cell is defined as:

$$V_i = \{x \in \mathcal{X} : \|x - p_i\| \leq \|x - p_j\| \text{ for all } j \neq i\}. \quad (32)$$

For a given Voronoi decomposition, the reconstruction noise is quantified as the expected squared distance from a random data point $X \sim \mu$ to its nearest generator:

$$E(k, P) = \mathbb{E}[\min_{i=1,\cdots,k} \|X - p_i\|^2] = \int_{\mathcal{X}} \min_{i=1,\cdots,k} \|X - p_i\|^2 \, d\mu(x). \quad (33)$$

For a fixed number of clusters $k$, the optimal generator configuration minimizes this noise:

$$E_k = \min_{P \subset \mathbb{R}^d, |P|=k} E(k, P). \quad (34)$$

To evaluate clustering quality, we introduce silhouette coefficient $S_k$, which measures both intra-cluster cohesion and inter-cluster separation. For any data point $x_i \in V_j$, the intra-class average Euclidean distance of $x_i$ and the minimum value of the average distance from $x_i$ to all Voronoi cells not belonging to itself are:

$$a_i = (\sum_{x_m \in V_j, m \neq i} \|x_i - x_m\|) / (|V_j| - 1), \quad (35)$$

$$b_i = \min_{\substack{l=1,2,\cdots,k \\ l \neq j}} (\sum_{x_m \in V_l} \|x_i - x_m\|) / |V_l|, \quad (36)$$

where $|V_j|$ is the total number of data points contained in the Voronoi cell $V_j$, $j = 1, 2, \cdots, k$.

The silhouette coefficient quantifies how similar a point is to its own cluster versus neighboring clusters, with values ranging from -1 to 1. A score close to 1 indicates high-quality clustering, while scores near -1 suggest points may be assigned to the wrong cluster. It for point $x_i$ is then defined as:

$$s_i = (b_i - a_i) / \max\{a_i, b_i\}. \quad (37)$$

The overall silhouette coefficient for the clustering is the average over all data points:

$$S_k = (\sum_{i=1}^{N} s_i) / N, \quad (38)$$

where $N = \sum_{j=1}^{k} |V_j|$, as $k$ increases, the silhouette coefficient $S_k$ typically reaches its maximum at a specific $k$ value, where this peak reflects the optimal balance between cluster cohesion and separation.

To balance these trade-offs, we introduce a regularized objective function that combines reconstruction noise with a complexity penalty and clustering quality:

$$J_k = E_k + \lambda_1 k - \lambda_2 S_k, \quad (39)$$

where $\lambda_1 > 0$ and $\lambda_2 > 0$ are regularization parameters controlling the trade-off between reconstruction accuracy, model complexity, and clustering quality.

According to quantization theory, for data distributions with smooth probability density $\mu$ on a compact support, the optimal quantization noise $E_k$ satisfies the following asymptotic relation:

$$\lim_{k \to \infty} k^{2/d} E_k = C(d, \mu), \quad (40)$$

where $d$ represents the data space dimensionality and $C(d, \mu)$ is a constant depending on both the dimension and data distribution. This leads to the approximation for sufficiently large $k$:

$$E_k \approx A k^{-2/d}, \quad (41)$$

where $A = C(d, \mu)$. Substituting this approximation into the regularized objective function yields:

$$J_k \approx A k^{-2/d} + \lambda_1 k - \lambda_2 S_k. \quad (42)$$

For typical data distributions, as illustrated in Fig. 11. $S_k$ is convex function with a maximum at a certain $k$ value, $d^2 S / dk^2 < 0$ near the optimum, confirming that $k^*$ is a local minimum of $J_k$.

This proof theoretically demonstrates that for any given dataset and regularization strength, there exists an optimal Voronoi cluster number $k^*$ that minimizes the combined reconstruction noise and model complexity. This provides the mathematical foundation for our choice of three-cluster decomposition in this work. Experimental results in Fig. 9 show that the three-cluster partitioning outperforms other partitioning schemes across all evaluation metrics, validating the consistency between theoretical analysis and experimental findings.

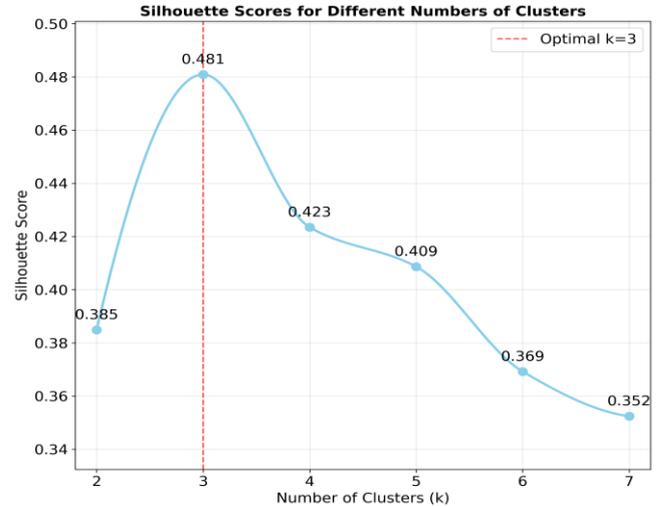

Fig. 12. Silhouette coefficient comparison for different Voronoi cluster numbers.

## V. CONCLUSION

Deep learning-driven CT reconstruction has advanced rapidly, but LDCT still faces challenges such as increased noise, compromised data fidelity, and underutilized global structural properties of raw pre-log projection data. This study proposes a novel diffusion Transformer framework for LDCT reconstruction, which operates in the pre-log domain and achieves regional decomposition via K-means clustering, this decomposition further segments sinograms into three clusters in the Voronoi space, constructs tailored multi-latent spaces to suppress aliasing, and integrates a diffusion model with data consistency constraints for noise reduction and structural preservation. Experiments on AAPM and CHAOS datasets show the framework outperforms baselines at noise levels, with fewer artifacts and better detail preservation, while ablation studies validate the synergy between K-means-Voronoi space decomposition and diffusion blocks. Given medical data constraints like cost and privacy issues, future work will adopt multi-domain and multi-model strategies to further improve LDCT reconstruction accuracy and advance structured modeling in the projection domain.

## ACKNOWLEDGMENTS

All authors declare that they have no known conflicts of interest in terms of competing financial interests or personal relationships.